\documentclass{article} % For LaTeX2e
\usepackage{iclr2025_conference,times}

\usepackage{amsmath,amsfonts,bm}

\def\eqref#1{equation~\ref{#1}}
\def\1{\bm{1}}

\DeclareMathAlphabet{\mathsfit}{\encodingdefault}{\sfdefault}{m}{sl}
\SetMathAlphabet{\mathsfit}{bold}{\encodingdefault}{\sfdefault}{bx}{n}

\makeatletter
\def\thanks#1{\protected@xdef\@thanks{\@thanks
        \protect\footnotetext{#1}}}
\makeatother

\usepackage{hyperref}       % hyperlinks
\usepackage{url}            % simple URL typesetting
\usepackage{booktabs}       % professional-quality tables
\usepackage{amsfonts}       % blackboard math symbols
\usepackage{nicefrac}       % compact symbols for 1/2, etc.
\usepackage{microtype}      % microtypography
\usepackage{xcolor}         % colors
\usepackage{multirow}
\usepackage{multicol}
\usepackage{graphicx}
\usepackage{enumitem}
\usepackage{amssymb}
\usepackage{bbding}
\usepackage{wrapfig}
\usepackage{subfigure}
\usepackage{capt-of}
\usepackage{authblk}
\usepackage{float}
\usepackage{xspace}

\usepackage{tipa}
\usepackage{pifont}
\usepackage{makecell}
\usepackage{utfsym}

\makeatletter
\DeclareRobustCommand\onedot{\futurelet\@let@token\@onedot}
\def\@onedot{\ifx\@let@token.\else.\null\fi\xspace}

\def\eg{\emph{e.g}\onedot} 
\def\ie{\emph{i.e}\onedot}

\def\wrt{w.r.t\onedot}

\makeatother

\title{MineWorld: a Real-Time and Open-Source Interactive World Model on Minecraft}

\author{
\textbf{Junliang Guo$^*$\thanks{$^*$Equal contribution. Correspondence to Junliang Guo.}, Yang Ye$^*$, Tianyu He$^*$, Haoyu Wu$^*$, Yushu Jiang, Tim Pearce, Jiang Bian}}
\affil{
Microsoft Research \\ 
{\footnotesize \texttt{\{junliangguo,v-yangye,tianyuhe, v-haoywu\}@microsoft.com}}
}
\affil{
\url{https://aka.ms/mineworld}
}

\iclrfinalcopy
\begin{document}

\maketitle

\begin{abstract}
World modeling is a crucial task for enabling intelligent agents to effectively interact with humans and operate in dynamic environments. In this work, we propose MineWorld, a real-time interactive world model on Minecraft, an open-ended sandbox game which has been utilized as a common testbed for world modeling. MineWorld is driven by a visual-action autoregressive Transformer, which takes paired game scenes and corresponding actions as input, and generates consequent new scenes following the actions. Specifically, by transforming visual game scenes and actions into discrete token ids with an image tokenizer and an action tokenizer correspondingly, we consist the model input with the concatenation of the two kinds of ids interleaved. The model is then trained with next token prediction to learn rich representations of game states as well as the conditions between states and actions simultaneously. In inference, we develop a novel parallel decoding algorithm that predicts the spatial redundant tokens in each frame at the same time, letting models in different scales generate $4$ to $7$ frames per second and enabling real-time interactions with game players.
In evaluation, we propose new metrics to assess not only visual quality but also the action following capacity when generating new scenes, which is crucial for a world model. Our comprehensive evaluation shows the efficacy of MineWorld, outperforming SoTA open-sourced diffusion based world models significantly. The code and model have been released.
\end{abstract}

\section{Introduction}
\label{sec:intro}

World models have been extensively studied for their potential ability to simulate and interact with various environments and actions taken by humans or agents~\citep{ha2018world,yang2023learning}. These models provide a computational framework that empowers intelligent agents to perceive surroundings, receive controls, and predict consequences. Thus, world models reduce reliance on real-world trials and automate complex tasks across industries, such as serving as a game engine~\citep{valevski2024diffusion,bruce2024genie,oasis2024,kanervisto2025world} or a planner in a reinforcement learning system~\citep{hafner2019dream,wu2024ivideogpt,agarwal2025cosmos}, illustrating their ability to improve decision-making, enable safe exploration, and facilitate scalable learning.

Recently, video generative models have shown a promising ability to learn commonsense knowledge from raw video datasets, ranging from physical laws in the real world~\citep{videoworldsimulators2024} to object interactions in games~\citep{parkerholder2024genie2,kanervisto2025world}, laying the foundation for their use as real-world simulators. 
However, foundamental challenges exist in ensuring the efficiency and controllability of these models, which are both crucial features for a desired world model. 

The efficiency bottleneck lies in the generation target of these video generative models, \ie, the latent videos representation encoded by visual tokenizers, consists of a large number of tokens (\eg, $40$k to $160$k tokens for $16$ frames with SoTA tokenizers~\citep{yang2024cogvideox,tang2024vidtok,wang2024vidtwin}). That leads to a substantial computational costs during inference, making real-time interactions with the model a major challenge.
Furthermore, the controllability requires the model to generate accurate outcomes based on a given control signal, which is challenging to evaluate due to the diverse nature of these signals. For instance, video generative models may be conditioned on textual descriptions~\citep{videoworldsimulators2024}, video demonstrations~\citep{zhang2025autoregressive}, and numeric features such as robotic arm movements~\citep{wu2024ivideogpt}, yet a standardized metric to quantify how well the generated results adhere to the input signals remains lacking.

In this work, we propose \textbf{MineWorld}, a real-time, open-source interactive world model on the game Minecraft. It is built upon an autoregressive Transformer, designed to overcome the challenges of the efficiency and controllability in video-based world modeling. MineWorld explicitly learns the correlation between visual states and control signals by tokenizing both game scenes and actions into discrete representations, which are concatenated interleaved as the input to the model.
To achieve real-time interactions between the model and humans\footnote{We provide the definition of real-time interaction for our world model in Section~\ref{sec:decoding}.}, we introduce a novel parallel decoding algorithm that significantly accelerates the autoregressive generation of Transformer. Instead of sequentially predicting each token step by step, our method exploits the dependencies between spatially adjacent tokens, allowing certain token groups to be predicted in parallel. This optimization results in over a more than $3\times$ speedup compared to standard autoregressive decoding, without sacreficing the quality of results. Equipped with this decoding algorithm, MineWorld is able to generate $4$ to $7$ frames in one second, making the real-time interaction between human and world model feasible.

In addition, to assess the controllability of MineWorld, we propose new evaluation metrics that extend beyond conventional video quality assessments. We first transform the actions in Minecraft to discrete tokens in one vocabulary. Then, after output videos are generated conditioned on previous frames and actions, we utilize an inverse dynamic model~(IDM)~\citep{baker2022video} to predict an action between consecutive generated frames. This predicted action can be treated as the executed action according to generated videos, while the input action is the ground truth one which serves as the condition. Therefore, the accuracy between the two kinds of actions reflect the controllability of this generative model. 

By addressing both efficiency and controllability, MineWorld advances the field of world modeling, offering the first high-quality and efficient framework for real-time simulation and interaction. To empower further research in this direction, we release our code and model weights. In summary, our contributions are threefold:
\begin{itemize}
    \item MineWorld, an open-sourced, real-time interactive world model powered by an autoregressive Transformer, establishes a new benchmark for world modeling.
    \item A novel parallel decoding algorithm that brings significant speedup in inference over standard autoregressive decoding, while maintaining high-quality generation results.
    \item New evaluation metrics for assessing the controllability, in order to validate whether the generated sequences faithfully adhere to control signals.
\end{itemize}

\section{Framework}
\label{sec:framework}

\begin{figure}[tbp]
    \centering
    \includegraphics[width=0.9\linewidth]{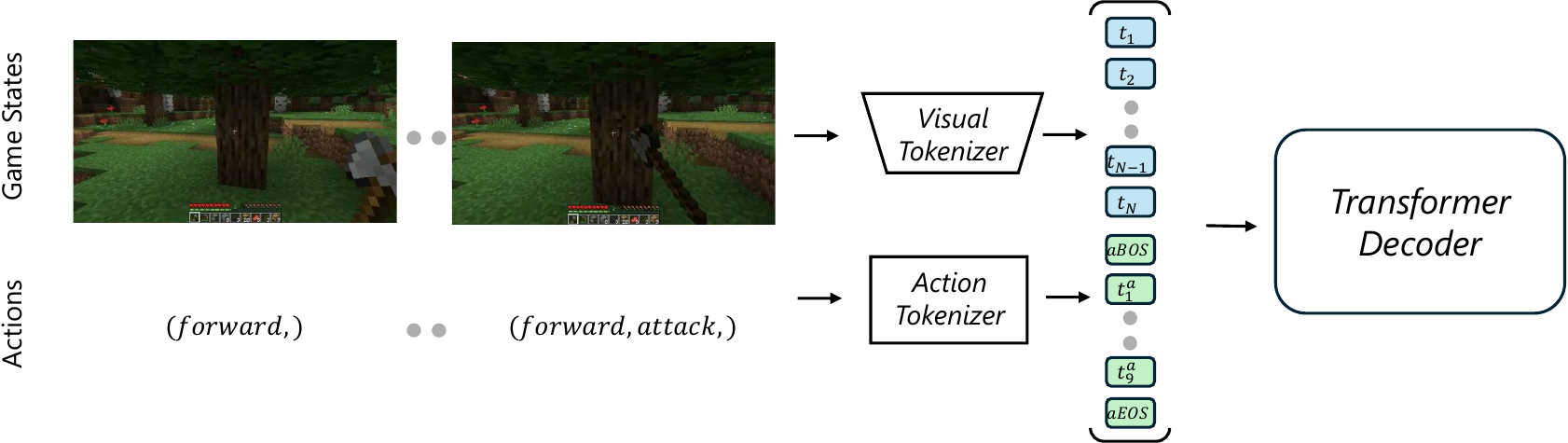}
    \caption{Illustrations of MineWorld model architecture. Visual and action tokenizers convert game states and actions into discrete tokens, which are concatenated and fed into a Transformer decoder as the input. The Transformer is then trained with an autoregressive objective.}
    \label{fig:arch}
\end{figure}

\subsection{Overview}
We will introduce the proposed MineWorld framework in this section. We start with the problem definition of the studied task. Denote $x_{i}$ as the visual game state from Minecraft at the $i$-th timestep, and $a_{i}$ as the action taken by users on the state, then the next game state $x_{i+1}$ represents the future when taking $a_i$ on $x_i$. The objective of our world model is to predict future game states based on past observations $x_{<i}$ and the current action $a_i$, modeling the following conditional distribution:
\begin{equation}
    \label{equ:world_model}
    p(x_{i+1} | x_{<i}, a_i).
\end{equation}

We build MineWorld on the autoregressive Transformer~\citep{vaswani2017attention}, considering its scaling behavior~\citep{kaplan2020scaling,brown2020language-gpt3} and good controllability (as will be shown in Section~\ref{sec:exp}). As illustrated in Figure~\ref{fig:arch}, the architecture consists of two modules, the tokenizers to convert videos and actions to discrete tokens, and a Transformer decoder to take the sequence of tokens as input and train in an autoregressive way. We introduce the details in the following.

\subsection{Architectures}
\label{sec:framework-arch}

\paragraph{Tokenizers}
The model input contains two different modalities, \ie, the game playing videos consist of sequences of states $x_i$, and actions consist of mouse and keyboard inputs. Therefore, we design different tokenizers to convert them into discrete tokens respectively. 

For game videos, we train a VQ-VAE~\citep{van2017neural,esser2021taming} as the visual tokenizer. Considering the interleaved manner of the game states and actions, we utilize an image-level (\ie, with spatial compressions) VQ tokenizer to convert each state to tokens independently, and leave the utilization of video-level tokenizers (\ie, with both spatial and temporal compressions)~\citep{tang2024vidtok} for future work. Specifically, we initialize the VQ tokenizer from from a public pre-trained checkpoint~\citep{patil2024amused} and then fine-tune it on the Minecraft dataset. The tokenizer has $16\times$ compression rates on both the height and width. As a result, for a clip of game video $x$ that contains $n$ states, the VQ tokenizer converts it into a sequence of quantized ids $t$, denoted as:
\begin{equation} 
\begin{aligned}
x &= (x_1, \cdots,x_n), \\
t &= (t_1, \cdots, t_{c}, t_{c+1}, \cdots, t_{2c}, t_{2c+1}, \cdots t_N),
\end{aligned}
\end{equation}
where $N=n \cdot c$ is the total length of the sequence, and $c$ is the number of ids to represent each state. 

The actions in Minecraft contains both continuous mouse movements that controls camera angles, and keyboard or mouse presses that represent discrete actions defined in the game engine such as \texttt{forward} and \texttt{attack}. To handle continuous movements, we follow previous practices~\citep{baker2022video} and quantize camera angles into discrete bins. For discrete actions, considering the mutual exclusive nature of certain action pairs (e.g., \texttt{forward} and \texttt{backward} cannot occur simultaneously), we categorize the actions into $7$ exclusive classes, each represented by a unique token. Additionally, we allocate $2$ tokens to encode camera angles and introduce special tokens \texttt{[aBOS]} and \texttt{[aEOS]} to mark the boundaries of an action sequence. As a result, each action is represented by a sequence of $11$ tokens, where each token corresponds to an action id from the complete action vocabulary. 
Dividing actions into exclusive classes not only helps reduce the sequence length of action tokens, but also makes it possible to validate the controllability of the model with classification based metrics, which will be introduced in Section~\ref{sec:eval}.

In conclusion, for each pair of game state and actions in the original input $(x_i, a_i)$, the tokenizers will transfer them into a flat sequence of discrete ids as:
\begin{equation}
    (t_{i*c+1}, \cdots, t_{(i+1)*c}, [\texttt{aBOS}], t^{a_i}_1, \cdots, t^{a_i}_9, \texttt{[aEOS]}).
\end{equation}

\paragraph{Transformer Decoder}
We design our Transformer following the LLaMA architecture~\citep{touvron2023llama} equipped with root mean square layer normalization~\citep{zhang2019root-rms} and Rotary Embeddings~\citep{su2024roformer-rope}. We train the model as a traditional autoregressive decoder, where each token is predicted conditioned on all previous tokens,
\begin{equation}
    \label{equ:training}
    f_{\theta}(t) = \prod_{i=1}^N p(t_i | t_{< i}).
\end{equation}
Note that we treat the tokens of game states and actions equivalently, allowing the model to learn their interleaved structure. As a result, the model jointly captures the conditional relationships between states and actions, enabling it to function as both a policy model (\ie, predicting actions based on previous observations $p(a_{i+1} | x_{<i+1})$)  and a world model (\ie, predict future states as described in Equ~(\ref{equ:world_model})) in inference. 
While our primary focus in this paper is on the model's performance as a world model, we also present case studies in the experiments to demonstrate its potential as a policy model.

\subsection{Parallel Decoding}
\label{sec:decoding}

\begin{figure}[tbp]
  \centering
   \includegraphics[width=1\linewidth]{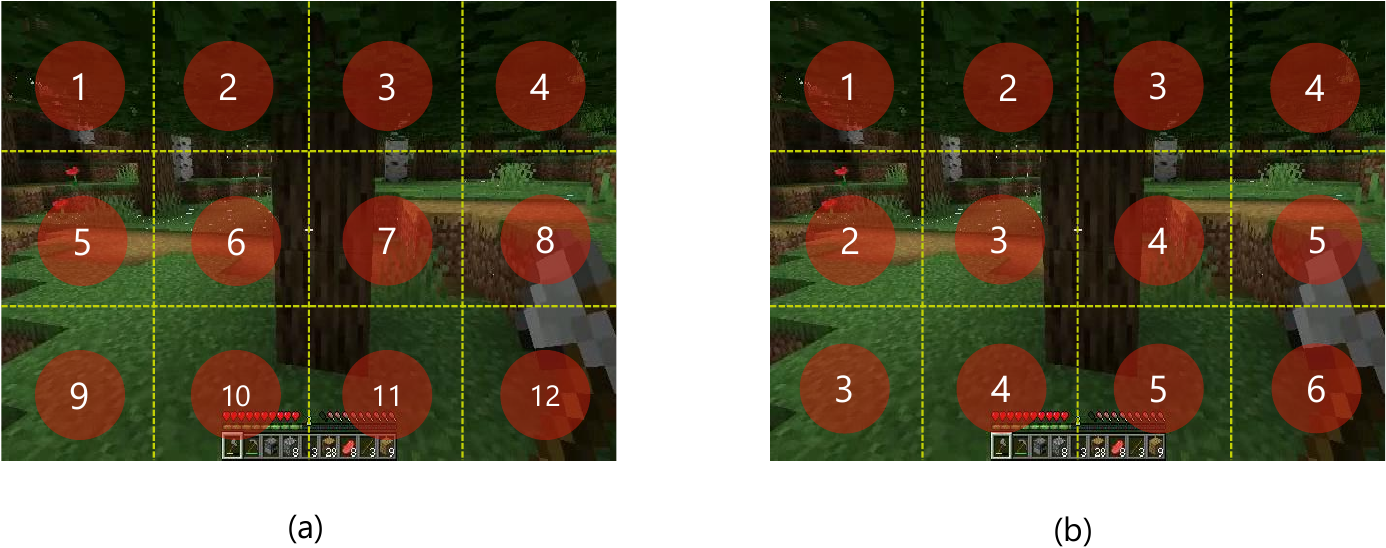}
   \caption{An illustration of two decoding algorithms. The game state is encoded into $12$ tokens. The number in each token represents that in which decoding iteration it is generated. (a) Autoregressive decoding, which follows a raster scanned order and generates each token sequentially. (b) Our proposed parallel decoding. For each generated token, the tokens in the adjacent rows and columns will be generated simutaneously in the next iteration.}
   \label{fig:decode-order}
\end{figure}

For a world model, it is crucial to generate real-time consequences \wrt the controling signals provided by the users. We provide the definition of ``real-time'' in our game scenario considering the Actions Per Minute~(APM)~\citep{wikipedia:apm} of professional players.
\paragraph{Real-Time Interactive World Model}
APM is a metric used to measure the number of in-game actions~(\eg, keypresses, mouse clicks) a player executes within one minute. Professional players in the game StarCraft typically have APMs around $250$ to $300$ in average, and $50$ for beginners~\citep{wikipedia:apm}. We also conduct an in-house test on amateur game players and find that they have $150$ APMs in average. Therefore, to develop a world model that achieves real-time interactions with users, it should generate more than $2$ Frames Per Second~(FPS) to keep up with amateur game players and $5$ FPS for professional ones.

While visual autoregressive models~\citep{liu2024elucidating, yu2023language, kondratyuk2023videopoet} can provide high-fidelity results for image and video generation tasks, their sequential nature poses a significant bottleneck on the efficiency when generating high-resolution, long-duration videos. Previous works~\citep{bai2024sequential, yu2023language} generate visual tokens following a fixed raster-scan order (\ie, from left-to-right and top-to-bottom) or random orders~\citep{chang2022maskgit,yu2024randomized}, without leveraging the inherent spatio and temporal dependencies in images and videos.

To facilitate the real-time interaction of our world model, we utilize Diagonal Decoding~\citep{ye2025fast}, a training-free parallel decoding algorithm by leveraging the spatial dependencies between adjacent image tokens, providing significant speedup compared to naive autoregressive decoding with negligible quality degradation. Specifically, our method processes tokens across different rows and columns simultaneously, as shown in Figure~\ref{fig:decode-order}. Let $x_{i,j}$ denote one of the generated tokens at row $i$ and column $j$ in the current game state, then in the next step, both tokens $x_{i,j+1}$ and $x_{i+1,j}$ will be generated at the same time. Denote the height and width of the patchified game state as $h$ and $w$, then the theoretical speedup ratio of our parallel decoding versus autoregressive decoding can be written as:
\begin{equation}
  r = \frac{h \times w}{h+w-1}.
  \label{eq:accelerate ratio}
\end{equation}
From the above equation, we can observe that the larger the image resolution, the faster our decoding speed. 

Our method utilizes the spatial redundency of adjacent tokens. However, due to the discrepency between training and inference, the speedup brings some performance degradation to the result. To solve the problem, we propose to fine-tune the autoregressively pre-trained model, by replacing the standard causal attention mask with the mask that aligns with our parallel decoding algorithm. As a result, our model is able to achieve real-time interaction frequency while maintaining good quality of generation results.

\paragraph{Discussion with Related Works}
Several prior works have explored parallel decoding in visual generative models. Lformer~\citep{li2023lformer} generate L-shaped token blocks in each iteration but requires training the model from scratch. More recently, ZipAR~\citep{he2024zipar} propose a parallel decoding algorithm for image generation by leveraging adjacent token dependencies. In contrast, our approach focuses on video generation, which presents additional challenges due to error accumulation across frames. Despite this increased complexity, our method achieves higher acceleration ratios relative to the autoregressive baseline, demonstrating the effectiveness of our parallel decoding strategy for real-time interactive applications.

\section{Evaluation}
\label{sec:eval}
To comprehensively evaluate the performance of world models in the Minecraft environment, we assess both the video quality and controllability quality of generation results. We introduce the construction and preprocessing of the dataset first.

\subsection{Dataset Construction and Preprocessing}
We utilize the VPT dataset~\citep{baker2022video} which consists of the pairs of recorded game playing videos and corresponding actions, where each frame is accompanied with the keyboard and mouse actions taken at the same time.
We filter out the frames with no recorded action and also the ones when GUI is open to eliminate its influence of results. We split long videos into short clips with $16$  frames considering the maximum context length of the model, and divide them into training, validation and test sets randomly. As a result, we train the model on $10$M video clips~(\ie, $160$M frames) and validate / test on $0.5$k / $1$k clips.

For each original video, we resize its resolution from $360\times 640$ to $224\times 384$, to reduce the computation cost while keeping the original aspect ratio for better visual quality. As introduced in Section~\ref{sec:framework-arch}, we adopt a VQ-VAE with $16\times$ spatial compression ratio and $8$k codebook size to transform each frame into $14\times 24$ patches, and finally a sequence of $336$ discrete tokens. In addtion to the $11$ action tokens encoded by our action tokenizer, we represent each pair of game state and action as $347$ tokens. As a result, for each training sample which contains $16$ pairs of them, we transform it into $5.5$k discrete tokens after preprocessing. And in total there is $55$B tokens in the training set.

\subsection{Evaluation Metrics}
\label{sec:eval-metric}

We utilize two types of metrics to assess the visual quality and the controllablility of the results generated by our model. For visual quality, we employ several common metrics including Fréchet Video Distance (FVD)~\citep{unterthiner2018towards}, Peak Signal-to-Noise Ratio (PSNR)~\citep{hore2010image}, Learned Perceptual Image Patch Similarity (LPIPS)~\citep{zhang2018unreasonable}, and Structural Similarity Index Measure (SSIM)~\citep{wang2004image}. 

Controllablility indicates to what extent the generated game states align with the provided actions and previous states, which is a crucial feature for an accurate world model. To evaluate it, we utilize an Inverse Dynamics Model~(IDM)~\citep{baker2022video} to infer the generated actions from the generated states and previous states. Specifically, the IDM is a bidirectional model which takes game states as input and predicts the actions between them. We utilize a highly accurate pre-trained IDM~\citep{baker2022video} which achieves $90.6 \%$ accuracy on keypresses.
However, due to the imbalanced nature of action labels (\eg, \texttt{forward} and \texttt{attack} appear much more frequently than \texttt{use} and \texttt{drop}) as well as the mixture of discrete and continuous actions (\eg, keyboard presses and camera movements), it is non-trival to design accurate and comprehensive evaluation metrics on the controllablility.
We propose two kinds of metrics to solve this problem.

\paragraph{Discrete Action Classification} 
Following the dividing policy in the action tokenizer, actions can be grouped into $9$ classes, where $7$ of them represent discrete action classes and the other $2$ represent camera movement angles. 
For discrete classes, each one of them contains two or three exclusive actions such as $(\texttt{forward}, \texttt{backward})$ and $(\texttt{left}, \texttt{right})$. We provide the full grouping results in Table~\ref{tab:subtask}. Then, by taking the provided action as the ground truth and the predicted action from IDM as the prediction, we can utilize commonly utilized classification metrics including precision, recall and F1 score to evaluate the classification accuracy. We report both the macro scores to reduce the effect of imbalanced labels.

\paragraph{Camera Movement.} 
Camera movement is represented by the player's rotational angle and handled separately from action prediction. Following the configuration in VPT~\citep{baker2022video}, we discretize the camera rotation along both the X and Y axes into 11 bins, each representing a specific range of angles. This design provides finer granularity for small movements and coarser bins for larger rotations, balancing precision and range. To evaluate the controllability of camera movements, we compute the L1 loss between the predicted and ground truth camera bins.

\subsection{Implementation Details}

For the visual tokenizer, we initialize it from a pre-trained checkpoint~\citep{patil2024amused} and then fine-tune it on the preprocessed VPT training data. We show the reconstruction results of visual tokenizers in Appendix~\ref{app:more_results}, and we can find that after fine-tuning, the tokenizer achieves good reconstrution quality on the validation set of VPT. 

For the Transformer decoder, we adopt LLaMA~\citep{touvron2023llama} architecture considering its advantageous to large-scale modeling. We experiment with different sets of hyper-parameters which result in models with $300$M, $700$M and $1.2$B parameters separately. 
The vocabulary size of the image tokenizer is $8192$, plus $70$ ids from the action vocabulary, and thus the final vocabulary size of our model is $8262$.
We use the Adam optimizer~\citep{kingma2014adam} with a cosine decay learning rate scheduler to train the model. The training is conducted on $32$ NVIDIA 40G A100 GPUs with PyTorch~\citep{paszke2019pytorch} for $200$k steps.
We introduce more training details in Appendix~\ref{app:model_config}.

\section{Experiments}
\label{sec:exp}

We present the results of MineWorld in this section. We compare our model with Oasis~\citep{oasis2024}, an open-sourced diffusion based world model on Minecraft.

\subsection{Main Results}

\begin{table}[tb]
\begin{center}
\small
\caption{Generation result of Oasis~\citep{oasis2024} and different scales of our models.``FPS'' indicates the number of frames generated per second by the model. ``P, R, F1'' denote the classification precision, recall and F1 scores on discrete actions. ``L1'' indicates the camera control loss.}
	\label{tab:main_results}
        \setlength\tabcolsep{6pt}
	\begin{tabular}{lc|c|cccc|cccc}
		\toprule[1pt]
		Method  & Param. & FPS$\uparrow$ & P$\uparrow$ & R$\uparrow$ & F1$\uparrow$ &  L1$\downarrow$ & FVD$\downarrow$  &  LPIPS$\downarrow$ & SSIM$\uparrow$ & PSNR$\uparrow$   \\
		\midrule
		Oasis & 500M & 2.58 & 0.49 & 0.44 & 0.41 & 2.60 & 377 & 0.53 & 0.36 & 14.38 \\ 
        \midrule
		\multirow{3}{*}{MineWorld} & 300M & \textbf{5.91} & 0.72 & 0.71 & 0.70 & 1.03 & 246 & 0.45 & 0.38 & 15.13  \\
		       & 700M & 3.18 & 0.72 & 0.71  & 0.70 & 1.04 & 231 & 0.44 & 0.38 & 15.32  \\
		       & 1.2B & 3.01 & \textbf{0.76} & \textbf{0.73} & \textbf{0.73} & \textbf{1.02}  &  \textbf{227} & \textbf{0.44} & \textbf{0.41} & \textbf{15.69}   \\
		\bottomrule[1pt]
	\end{tabular}
 \vspace{-3mm}
\end{center}
\end{table}

We provide the results of our models and the Oasis baseline in Table~\ref{tab:main_results}. Both the video quality and controllability accuracy scores are presented. 
We use the open-sourced $500$M model and inference code from Oasis to obtain its results. Compare with it, all of our models achieve better results on both aspects of evaluation metrics. Note that while they claim good video quality and fast inference speed in their blog~\footnote{\url{https://oasis-model.github.io}}, it requires a larger model equipped with dedicated chips and inference engine, which are public inaccessible. On the other hand, our proposed parallel decoding algorithm is orthogonal to hardware and system level optimizations, and thus have the potential to achieve faster inference speed if strengths from both sides are combined.

From the comparisons between different model scales of MineWorld, we observe a clear scaling behaviour. Larger models achieve better performance on both the controllability and video quality. As for the inference speed, thanks to the proposed parallel decoding algorithm, even the largest $1.2$B model reaches a FPS of $3$, \ie, generating $3$ game states in one second and thus can respond to $180$ Actions Per Minute~(APM), an interaction frequence for most amateur game players. The most efficient $300$M model achieves an APM around $360$, which is able to interact with top professional players.

\subsection{Analysis}
\label{sec:analysis}

\paragraph{The Effect of Parallel Decoding}

\begin{table}[tb]
\begin{center}
\small
\caption{The performance of different scales of MineWorld models with default autoregressive decoding and the proposed parallel decoding. ``w/'' or ``w/o FT'' indicate results after fine-tuning the autoregressive pre-trained model with the parallel attenion mask or not. We choose several main metrics on the efficiency and effectiveness to make the table compact.}
	\label{tab:parallel}
        \setlength\tabcolsep{10pt}
	\begin{tabular}{lc|cccc}
		\toprule[1pt]
		Param. & Decoding & FPS$\uparrow$ & F1$\uparrow$ & PSNR$\uparrow$ & FVD$\downarrow$ \\
		\midrule
		\multirow{3}{*}{300M} & AT & 2.00 & 0.70 & 15.63 & 223  \\
        	                 & Parallel w/ FT & 5.91 & 0.70 & 15.13 & 246   \\
        	                 & Parallel w/o FT & 5.91 & 0.69 & 14.98 &  275 \\
        \midrule
		\multirow{3}{*}{700M} & AT  & 1.08 & 0.73  & 15.74 & 210 \\
        	                 & Parallel w/ FT & 3.18 &  0.71  & 15.32 & 231  \\
        	                 & Parallel w/o FT & 3.18 & 0.70 & 15.27 & 247 \\ 
        \midrule
		\multirow{3}{*}{1.2B} & AT & 0.89 & 0.72 & 16.06 &  203 \\
        	                 & Parallel w/ FT & 3.01 & 0.73 & 15.69 & 227  \\
        	                 & Parallel w/o FT & 3.01 & 0.70 & 15.30 & 258  \\
		\bottomrule[1pt]
	\end{tabular}
 \vspace{-3mm}
\end{center}
\end{table}

We compare the performance of autoregressive decoding and the proposed parallel decoding in Table~\ref{tab:parallel}. The conclusions are twofold. Firstly, larger models perform better with parallel decoding even without fine-tuning. For example, the $1.2$B MineWorld model achieves a $3\times$ speedup in inference latency with parallel decoding, while preserving comparable performance in both controllability and video quality relative to autoregressive decoding.

Secondly, fine-tuning with the parallel attention mask consistently enhances the performance of parallel decoding. For smaller models, such as the $300$M variant, fine-tuning enables the model to match the autoregressive baseline, while still benefiting from a $3\times$ speedup in inference latency. These results suggest an exciting direction for future research: training the model from scratch with the parallel attention mask may lead to faster convergence and lower overall training costs.

\paragraph{The Metrics on Controllability}

\begin{wraptable}{r}{0.35\textwidth}
    \centering    
    \caption{The correlation between the proposed metrics and human evaluation on controllability.} 
    \begin{tabular}{l|c}
    \toprule
      F1  & $0.81$ \\    
      Human  & $4.21$ \\
      $r$ & $0.56$ \\
      p-value & $0.01$ \\
    \bottomrule  
    \end{tabular}       
    \label{tab:correlation}
\end{wraptable}

To validate that the metrics proposed in Section~\ref{sec:eval-metric} accurately measure the controllability of the model, we conduct human evaluation and calculate the correlation coefficient between the proposed metrics and humane evaluation. Specifically, we sample $20$ game video clips from the test set, and invite $5$ experienced game players to score each video clip from the aspect of action following. Each participant was presented with the pair of game states and their corresponding actions, and asked to rate each sample on a scale of 1 to 5. We calculated the average score as the final result. The pearson correlation coefficient is calculated over all samples.

We conduct experiments on the $700$M model with autoregressive decoding, and list the results in Table~\ref{tab:correlation}, where $r$ represents the pearson correlation coefficient and p-value represents the corresponding probability value. As a result, the classification based evaluation metric has a significant positive correlation with human evaluation, showing the correctness of the proposed metrics.

\subsection{Case Study}

\begin{figure}[tb]
  \centering
  \includegraphics[width=1.0\linewidth]{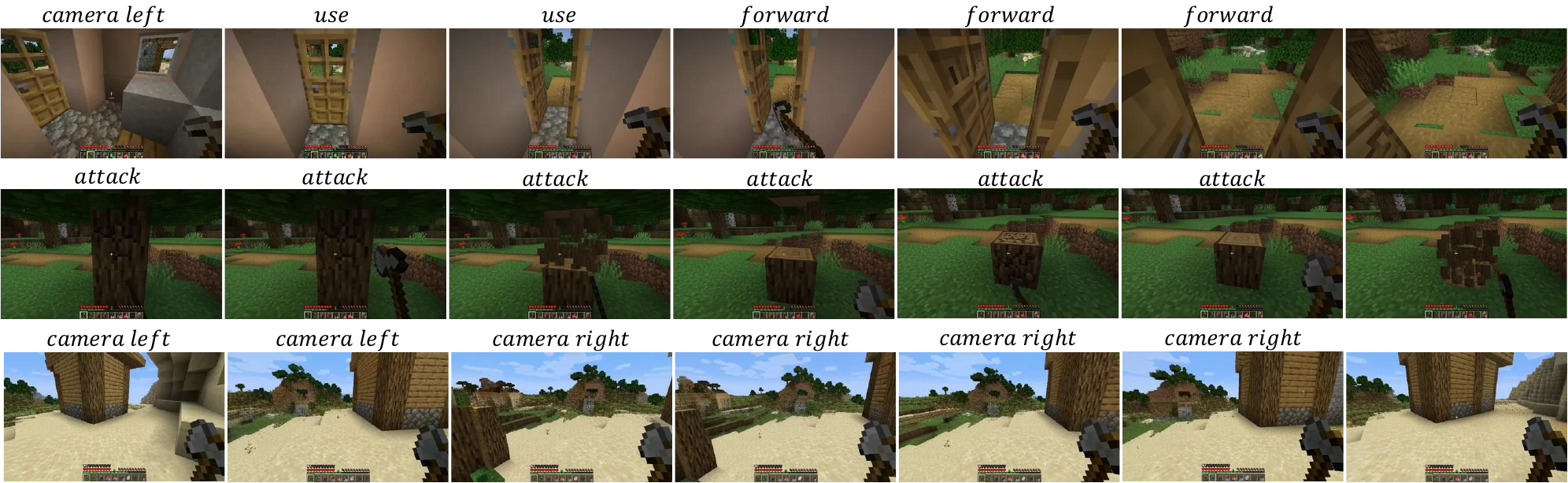}
  \vspace{-3mm}
  \caption{Case study on MineWorld $700$M model. The first game state and actions in following steps are provided as input, based on which the model generates consequent game states. For more cases and videos, please visit our project page.}
  \label{fig:case}
\end{figure}

We provide case studies on MineWorld in this section, to demonstrate the controllability and video quality of the model. All results are generated by the $700$M model with autoregressive decoding.

\paragraph{General Capability}
We show several general capability of MineWorld in Figure~\ref{fig:case}, where the initial game state is provided, and the model generates subsequent states based on corresponding actions. In the first example, the actions guide the model to open a door and walk outside. The model successfully generates the door-opening process and accurately renders the unseen outdoor environment. 
The second example depicts the process of chopping wood, where the model captures fine-grained details, such as the wood’s cross-section and the explosion effect when the chop is complete.
In the third case, a house appears in the initial frame, and the camera first pans left and then returns to the right. The model correctly regenerates the same house with detailed fidelity when the camera returns.

These cases show that MineWorld learns foundamental physical knowledge of Minecraft, and can generate accurate reactions when receiving diverse actions. Besides, the strong video generation capacity empowers the model generates high-fidelity, coherent and consistent video results.

\paragraph{Controllability}

\begin{figure}[tb]
  \centering
  \includegraphics[width=1.0\linewidth]{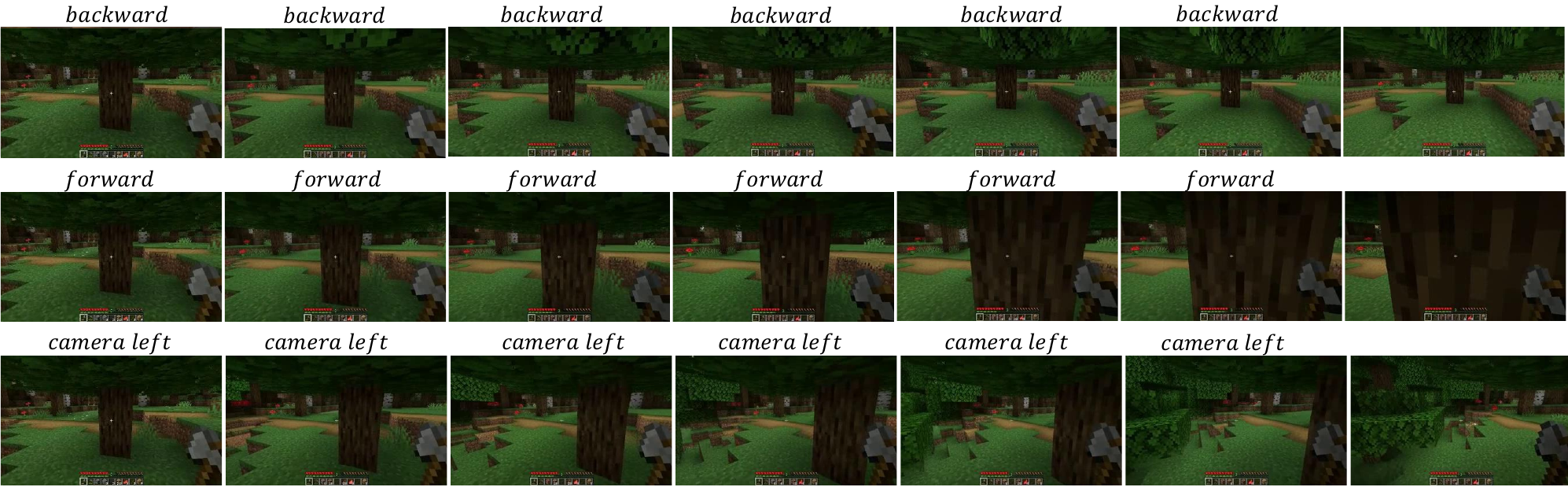}
  \vspace{-3mm}
  \caption{Case study on the controllability of MineWorld. Providing the same initial game state and different actions, the model generates different results correspondingly.}
  \label{fig:case_control}
\end{figure}

We illustrate the controllability of MineWorld in Figure~\ref{fig:case_control}, which shows the different generation results start from the same game state but with different actions. As a result, MineWorld accurately generates responses w.r.t each action.

\paragraph{Serve as an Agent}

\begin{figure}[tb]
  \centering
  \includegraphics[width=1.0\linewidth]{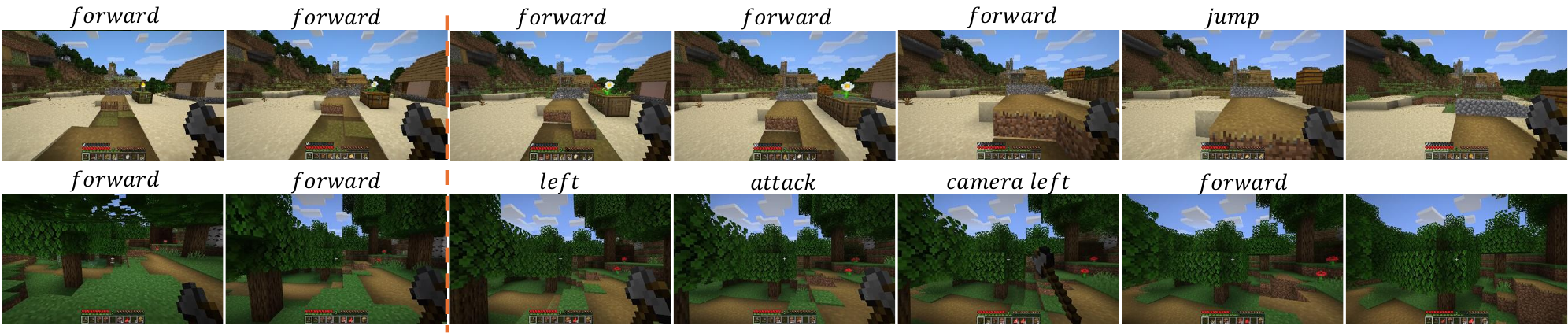}
  \vspace{-3mm}
  \caption{Case study on MineWorld as a gaming playing agent. By providing several initial game states and actions~(splited by the dashed line), MineWorld continues to play the game by itself through generating future game states and actions in an iterative manner.}
  \label{fig:case_dream}
\end{figure}

Since MineWorld predicts both game state tokens and action tokens during training, it naturally acquires the capabilities of both a world model and a policy model. This dual functionality enables MineWorld to serve as a self-contained game agent, capable of playing the game autonomously. Specifically, given a few initial game states and actions provided by users, the model can iteratively predict future states and actions, effectively simulating the long-horizon gameplay.

As shown in Figure~\ref{fig:case_dream}, MineWorld generates diverse and contextually appropriate actions, as well as accurate, high-fidelity game states. This demonstrates its potential not only as an interactive world simulator but also as a foundation for developing game-playing agents. 

\section{Conclusion and Limitations}

We present MineWorld, the first open-source, real-time interactive world model for Minecraft. Using an open dataset of game states and corresponding actions, we tokenize both modalities with separate tokenizers and train an autoregressive Transformer decoder via next-token prediction, using the interleaved sequence of state and action tokens as input. To support real-time interaction, we exploit the redundancy between adjacent image tokens and introduce a parallel decoding algorithm, achieving consistent speedups over standard autoregressive decoding without sacrifacing the generation quality. Through our comprehensive evaluation pipeline, we show that MineWorld achieves not only the strong controllability and video quality, but the fast inference speed of $2\sim6$ FPS which enables real-time interaction with professional game players.

\paragraph{Limitations}
MineWorld is trained exclusively on Minecraft data at a fixed, downsampled resolution, limiting its ability to generalize to other video domains (\eg, internet videos) or generate outputs at higher resolutions. The downsampling process may also lead to the loss of fine-grained details in game states. The maximum input length of the model is set to $5.5$k tokens, corresponding to $16$ state-action pairs. While MineWorld demonstrates strong temporal consistency within this range, it is not guaranteed when the distance between game states exceeds this range. 

\bibliography{main}
\bibliographystyle{iclr2025_conference}

\newpage
\appendix

\section{Minecraft Action Space}

\subsection{Details of Action Following Metric}

\begin{table}[htbp]
\centering
\caption{Classification Tasks and Their Labels}
\begin{tabular}{ l|c|c }
\toprule[1pt]
\textbf{Task Type} & \textbf{Actions} & \textbf{Labels} \\
\midrule
\multirow{3}{*}{Triple Classification} & \texttt{forward}, \texttt{backward} & \texttt{forward}, \texttt{backward}, \texttt{null} \\
& \texttt{left}, \texttt{right} & \texttt{left}, \texttt{right}, \texttt{null} \\
& \texttt{sprint}, \texttt{sneak} & \texttt{sprint}, \texttt{sneak}, \texttt{null} \\
\midrule
\multirow{4}{*}{Binary Classification} & \texttt{use} & \texttt{use}, \texttt{null} \\
& \texttt{attack} & \texttt{attack}, \texttt{null} \\
& \texttt{jump} & \texttt{jump}, \texttt{null} \\
& \texttt{drop} & \texttt{drop}, \texttt{null} \\
\bottomrule
\end{tabular}
\label{tab:subtask}
\end{table}

The action space in Minecraft is inherently complex. To effectively validate the quality of action execution, we simplify the scenario by focusing on the 10 most common actions and camera movements.
In addition, treating action prediction as a simple multi-class classification task does not fully capture the complexity of predicting sequential actions. To address this, we divide the $10$ actions into $3$ triple classification tasks and $4$ binary classification tasks, based on the exclusion relationships between the actions. In cases where the IDM model predicts two mutually exclusive actions at the same frame, we classify it as "no action" in line with common game behavior. Finally, we compute average macro precision, recall and F1 scores over all tasks.

\subsection{Task-Specific Action Accuracy}

\begin{table}[htbp]
\centering
\caption{Precision, Recall, and F1 scores for the 700 MineWorld model across different tasks.}
\begin{tabular}{ l|c|c|c|c|c|c|c }
\toprule[1pt]
\textbf{Metric} & \textbf{Forward-Backward} & \textbf{Left-Right} & \textbf{Sprint-Sneak} & \textbf{Use} & \textbf{Attack} & \textbf{Jump} & \textbf{Drop} \\
\midrule
\textbf{Precision} & 0.807 & 0.808 & 0.765 & 0.742 & 0.729 & 0.760 & 0.500 \\
\textbf{Recall} & 0.773 & 0.745 & 0.788 & 0.648 & 0.749 & 0.847 & 0.498 \\
\textbf{F1} & 0.782 & 0.771 & 0.750 & 0.682 & 0.736 & 0.796 & 0.499 \\
\bottomrule
\end{tabular}
\label{tab:specific_prf}
\end{table}

After defining the sub-classification tasks in Table \ref{tab:subtask}, we can analyze the model’s performance in more detail with respect to action prediction. The table \ref{tab:specific_prf} below shows the precision, recall, and F1 score for each task in our 700M MineWorld model. The results indicate that the "drop" action performs significantly worse than the other actions, suggesting that it is more challenging for the model to learn.

\section{Model Configurations}
\label{app:model_config}

\begin{table}[htbp]
    \centering
    \caption{The configuration of different size of models.}
    \begin{tabular}{l|c|c|c|c}
        \toprule[1pt]
            & \textbf{Hidden dim} & \textbf{MLP dim} & \textbf{Num. Heads} & \textbf{Num. Layers} \\
        \midrule
        300M &  1024 & 4096 & 16 & 20  \\
        700M &  2048 & 4096 & 32 & 20 \\
        1.2B &  2048 & 8192 & 32 & 20 \\
        \bottomrule[1pt]
    \end{tabular}
    \label{tab:model_arch}
\end{table}

\begin{table}[htbp]
    \centering
    \caption{Optimization hyperparameters.}
    \begin{tabular}{c|c}
        \toprule[1pt]
       \textbf{Hyperparameter}  &  \textbf{Value} \\
       \midrule
       Learning rate scheduler   &  cosine \\
       Learning rate & $3e^{-4}$ \\
       Warm up steps & 10000 \\
       Weight decay & 0.1 \\
       Optimizer & AdamW \\
       AdamW betas & $(0.9, 0.95)$ \\
       Maximum Positions & $5376$ \\
       \bottomrule[1pt]
    \end{tabular}
    \label{tab:hyperparams}
\end{table}

To validate the scaling behavior of the Transformer decoder, we train three different sizes of the model within the LLaMA architecture: $300$M, $700$M, and $1.2$B. We tune the hidden dimension, intermediate dimension, and the number of layers to achieve different model sizes. The configuration of these models are listed in Table~\ref{tab:model_arch}. The hyperparameters of the optimizer used to train the model are listed in Table~\ref{tab:hyperparams}.

\section{More Results}
\label{app:more_results}

\subsection{Reconstruction Results of Visual Tokenizer}

\begin{table}[htbp]
\begin{center}
\footnotesize
\caption{The reconstruction performance of the visual tokenizers on the validation set.}
	\label{tab:visual_tokenizer}
        \setlength\tabcolsep{8pt}
        \renewcommand\arraystretch{1.1}
	\begin{tabular}{l|cccc}
		\toprule[1pt]
		Visual Tokenizer & PSNR$\uparrow$ & SSIM$\uparrow$ & LPIPS$\downarrow$ & rFID$\downarrow$ \\
		\midrule
		Amused~\citep{patil2024amused} & 25.91 & 0.758 & 0.238 & 35.05 \\
        Ours & 29.24 & 0.816 & 0.134 & 18.93 \\
		\bottomrule[1pt]
	\end{tabular}
\end{center}
\end{table}

We evaluate the reconstruction performance of the pre-trained Amused VQ-VAE~\citep{patil2024amused} and the one after fine-tuned on the pre-processed VPT data. After fine-tuning, the performance is significantly improved, showing the necessity of this step.

\end{document}